\documentclass[10pt, a4paper]{article}
\usepackage{lrec}
\usepackage{graphicx}
\usepackage{tabularx}
\usepackage{soul}
\usepackage{multirow}

\usepackage{epstopdf}
\usepackage[latin1]{inputenc}

\usepackage{hyperref}
\usepackage{xstring}

\usepackage{subcaption}
\usepackage{breakcites}

\title{Face2Text: Collecting an Annotated Image Description Corpus for the Generation of Rich Face Descriptions}

\name{Albert Gatt$^{\ast}$, Marc Tanti$^{\ast}$, Adrian Muscat$^{\ast}$, Patrizia Paggio$^{\ast}$$^{\dagger}$, Reuben A Farrugia$^{\ast}$,  \\ {\bf \large Claudia Borg$^{\ast}$, Kenneth P Camilleri$^{\ast}$, Michael Rosner$^{\ast}$, Lonneke van der Plas$^{\ast}$}}

 \address{$^{\ast}$University of Malta \\
         {\{albert.gatt, marc.tanti, adrian.muscat, patrizia.paggio, reuben.farrugia\}@um.edu.mt} \\
         {\{claudia.borg, kenneth.camilleri, mike.rosner, lonneke.vanderplas\}@um.edu.mt}
          \\
          $^{\dagger}$University of Copenhagen \\
          paggio@hum.ku.dk
   }

\abstract{
The past few years have witnessed renewed interest in NLP tasks at the interface between vision and language. One intensively-studied problem is that of automatically generating text from images. In this paper, we extend this problem to the more specific domain of face description. Unlike scene descriptions, face descriptions are more fine-grained and rely on attributes extracted from the image, rather than objects and relations. Given that no data exists for this task, we present an ongoing crowdsourcing study to collect a corpus of descriptions of face images taken `in the wild'. To gain a better understanding of the variation we find in face description and the possible issues that this may raise, we also conducted an annotation study on a subset of the corpus. Primarily, we found descriptions to refer to a mixture of attributes, not only physical, but also emotional and inferential, which is bound to create further challenges for current image-to-text methods.
 }

\begin{document}

\maketitleabstract

\section{Introduction}

\begin{figure*}[t!]
	\centering
	\begin{subfigure}{0.45\textwidth}
      \caption{Male example
          \label{fig:example_male}
      }
      \centering
      \includegraphics[width=0.4\textwidth]{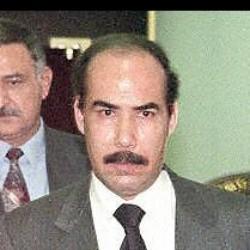}
      \begin{itemize}
        \item I see a serious man. Such facial expressions indicate that the man is highly committed and dedicated to his work
        \item A middle eastern gentleman struggling with an administrative problem
        \item criminal
        \item Longish face, receding hairline although the rest is carefully combed with a low parting on the person's left. Groomed mustache. Could be middle-eastern or from the Arab world. Double chin and an unhappy face. Very serious looking.
      \end{itemize}
	\end{subfigure}
	\hfill
    %
	\begin{subfigure}{0.45\textwidth}
      \caption{Female example
          \label{fig:example_female}
      }
      \centering
      \includegraphics[width=0.4\textwidth]{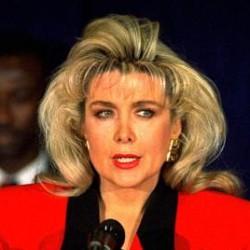}
      \begin{itemize}
        \item blonde hair, round face,  thin long nose
        \item While female , American stylish blonde hair and blue or green eyes wearing a suit , public speaks person
        \item Middle aged women, blond (natural ?) well groomed (maybe over groomed). Seems to be defending/justifying herself to a crowd/audience. Face of remorse/regret of something she has done.
        \item An attractive woman with a lovely blonde hair style, she looks pretty seductive with her red lips. She looks like a fashion queen for her age.
      \end{itemize}
    \end{subfigure}
	\caption{
		\label{fig:example}
		Examples of descriptions collected for two faces.
	}
\end{figure*}

This paper describes an annotation project that is being conducted by a cross-disciplinary group of researchers at the University of Malta, the RIVAL (Research In Vision And Language) group, to create a corpus of human face images annotated with rich textual descriptions. The initial goal of the project was to investigate in general how users describe images of human faces, and ultimately to create a resource that could be used to train a system to generate descriptions potentially as varied and rich as possible, thus moving the state-of-the-art of automatic description generation for images of faces from a feature-based process to one that takes advantage of complex textual descriptions. Here we report on a preliminary version of the corpus, focussing on how it was collected and evaluated.\footnote{The corpus can be downloaded by visiting \url{https://github.com/mtanti/face2text-dataset/}.}

\section{Background}

Automatic image description research can rely on a wide range of  image-description datasets. Such datasets consist of images depicting various objects and actions, and associated descriptions, typically collected through crowd-sourcing. The descriptions verbalise the objects and events or relations shown in the images with different degrees of granularity. For example, the most widely-used image captioning datsets, such as Flickr8k \cite{hodosh2013}, Flickr30k \cite{young2014}, VLT2K \cite{ElliottKeller2013}, and MS COCO \cite{MSCOCO}, contain images of familiar scenes, and the descriptions are restricted to the 
`concrete  conceptual' level \cite{hodosh2013}, mentioning what is visible, while minimising inferences that can be drawn from the visual information. Other datasets are somewhat more specialised. For example, the Caltech-UCSD Birds and Oxford Flowers-102 contain fine-grained visual descriptions
of images of birds and flowers respectively 
\cite{reed2016}. Some datasets also contain captions
 in different languages, as in the case of Multi30k \cite{multi30k}. A more extensive overview of image caption datasets can be found in \newcite{Bernardietal2015}. Although images of faces may be included in these datasets, none of them specifically targets face descriptions.

There are several datasets of faces that are widely used by the image processing community including the LFW \cite{LFW,Learned-Miller2016}, MegaFace \cite{kemelmacher2016megaface} and IJB-C datasets \cite{Klare2015}.
These datasets however do not have labelled attributes.
The LFWA \cite{LFW,Learned-Miller2016} and CelebA \cite{liu2015faceattributes} datasets on the other hand contain
images that are labelled with features mainly referring not only to physical facial attributes, such as skin colour and hair style, but also attributes of the person, e.g. age and gender. A number of datasets also focus specifically on emotion recognition and rendering from images of faces \cite{yin2006,NimStim}. The expressions are typically either acquired after specific emotions were elicited in the photographed subjects or posed by actors, and were subsequently validated by asking annotators to tag each image with emotion labels. None of these datasets pair images with text beyond simple feature labels.

The Face2Text dataset, a preliminary version of which is described in this paper, aims to combine characteristics from several existing datasets in a novel way by re-using a collection of images of human faces collected {\em in the wild}, in order to collect rich textual descriptions
of varying semantic granularity and syntactic complexity, which refer to physical attributes, emotions, as well as inferred elements which are not necessarily directly observable in the image itself.

There are a number of reasons why face descriptions are an interesting domain for vision-language research. Descriptions of faces are frequent in human communication, for example when one seeks to identify an individual or distinguish them from another person. They are also pervasive in descriptive or narrative text. The ability to adequately describe a person's facial features can give rise to more humanlike communication in artificial agents, with potential applications to conversational agents and interactive narrative generation, as well as forensic applications in which faces need to be identified from textual or spoken descriptions. Hence, the primary motivation for collecting the Face2Text corpus is that it promises to provide a resource that can break new ground in the problem of automatic description
and retrieval 
of images beyond the current focus on naturalistic images of scenes and common objects.

In corpora which target  a specific domain (for example, birds, plants, or human faces), the descriptions solicited tend to be fairly detailed and nuanced. This raises the possibility of more in-depth investigation of the use of language in a specific domain, including the identification and description of salient or distinguishing features (say, eye colour, the shape of a face, emotion or ethnicity) and the investigation of the conditions under which they tend to feature in human descriptions.
Indeed, descriptions of faces produced by humans are often {\em feature-based}, focussing on distinguishing physical characteristics of a face and/or transient emotional states. Alternatively, they may involve inference or analogy. Examples of such descriptions can be seen in Figure \ref{fig:example}.

Nevertheless, assuming the existence of an appropriate dataset, architectures for generating face descriptions are likely to share many of the challenges in the more familiar image description task. Hence, it is worth briefly outlining the ways in which the latter has been approached.
Approaches to image description generation are based either on caption retrieval or direct generation \cite{Bernardietal2015}.

In the generation-by-retrieval approach, human authored descriptions for similar images are stored in a database of image-description pairs. Given an input image that is to be described, the database is queried to find the most similar images to the input image and the descriptions of these images are returned. The descriptions are then either copied directly (which assumes that descriptions can be reused as-is with similar images) or synthesized from extracted phrases. There are examples of retrieval in visual space \cite{Ordonez2011,Kuznetsova2012} and of retrieval in multimodal space \cite{hodosh2013,Socher2014}.

On the other hand, direct generation approaches attempt to generate novel descriptions using natural language generation techniques. Traditionally, this was achieved by using computer vision (CV) detectors which are applied to generate a list of image content (e.g objects and their attributes, spatial relationships, and actions). These are fed into a classical NLG pipeline that produces a textual description, verbalising the salient aspects of the image \cite{Kulkarni2011,Mitchell2012}. The state of the art in image description makes use of deep learning approaches, usually relying on a neural language model to generate descriptions based on image analysis conducted via a pre-trained convolutional network \cite{Vinyals2015,Mao2015,Xu2015,Rennie2016}. While these systems are currently the state of the art, they suffer from a tendency to generate repetitive descriptions by generating a significant amount of descriptions that can be found as-is in the training set \cite{Devlin2015,Tanti2018}. This suggests that the datasets on which they are trained are very repetitive and lack diversity.

State of the art image captioning requires large datasets for training and testing. While such datasets do exist for scene descriptions, no data is currently available for the face description task, despite the existence of annotated image datasets. In the following section we describe how we addressed this lacuna, initiating an ongoing crowd-sourcing exercise to create a large dataset of face descriptions, paired with images which are annotated with physical features. Our long-term aim is to extend the scope of current image-to-text technology to finer-grained, attribute-focussed descriptions of specific types of entities, akin to those of birds and flowers \cite{reed2016}, rather than objects and relations in scenes.
\begin{figure*}[t]
	\centering
	  \includegraphics[width=0.75\textwidth]{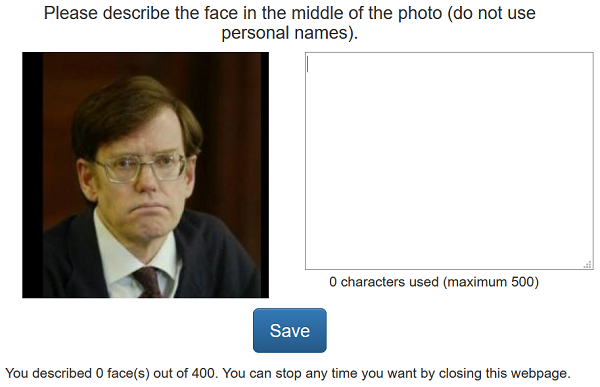}
      \caption{
          \label{fig:website_screenshot}
          Screen shot of the crowd sourcing website asking a visitor for a description.
      }
\end{figure*}

\section{The Face2Text dataset}
Descriptions for the Face2Text dataset are currently being collected via a custom-made crowd-sourcing website. A screen shot of the web page requesting a description of a face is shown in Figure~\ref{fig:website_screenshot}.

%
\paragraph{Pilot study} Prior to launching the crowd-sourcing interface, we conducted a small pilot study among approximately ten individuals, who were asked to write descriptions of a sample of the faces in the dataset. These were then used to provide participants with some examples of what was expected of them (see below).

\paragraph{Data} 400 Images were selected\footnote{http://rival.research.um.edu.mt/facedesc/} from the Faces in The Wild dataset\footnote{http://tamaraberg.com/faceDataset/} \cite{berg2005}, with the aim of collecting as many descriptions as possible for each image. These are close-up images of the faces of public personalities, taken in naturalistic contexts (that is, without controlling for background, or other `noise' in the image). The selected images were evenly divided between pictures of males and females.

\paragraph{Procedure} The system is designed to allow a participant to describe as many images as they wish. At any stage in the process, a participant is shown an image selected randomly among those that have been given the fewest descriptions so far. In the long-term this equalises the number of descriptions per image.

Participants were given minimal instructions in order to encourage as much variation as possible.
This was done because the precise nature of face descriptions, and the distribution of various features, is, to our knowledge, under-researched. Hence, rather than explicit instructions regarding what they were meant to write, participants were shown four examples of faces, each accompanied by three different descriptions collected through our preliminary pilot study.

The main points in the instructions that were given are the following:

\begin{itemize}
\item No recognisable faces are to be identified by name.
\item Descriptions should not be too short, but should not exceed a maximum of 500 characters.
\item Only the face in the middle of the image should be described (some of the images include multiple faces which happened to be near the target face).
\end{itemize}

After reading the instructions, participants were requested to enter basic demographic information, specifically: gender, age bracket, country of origin and proficiency in English. The latter is a self-rated forced choice among the options {\em native speaker, non-native but fluent} and {\em not fluent}. Only one respondent has rated themselves as non-fluent.

Participants could interrupt the study at any point. However, the system saved session variables, meaning that there was a limited time period during which participants could revisit the online interface and resume the description exercise from where they had left off.

\paragraph{Participation and dataset} Participation was voluntary and no financial or other incentives were offered to participants. To date, the crowdsourcing experiment has been advertised among staff and students in the University of Malta as well as on social media. A total of 1,400 descriptions have been collected from 185 participants. All 400 images have been described at least 3 times, with approximately 270 images having 4 descriptions.

\begin{table*}[!ht]
\begin{tabular}{|p{5cm}|c|l|p{3cm}|}
\hline
{\bf Question} & {\bf Positive responses} & {\bf Overall agreement (Fleiss's $\kappa$)} & {\bf Average $\kappa$ agreement with control annotator}\\ 
\hline
1. Is the description garbage? & 13\% & 0.9 & 0.65 (0.15)\\
2. Does the description contain elements which are inferred but external to the image? & 46\% & 0.48 & 0.39 (0.16)\\
3. Does the description refer to the emotional state of the face? & 44\% & 0.87 & 0.49 (0.33)\\
4. Does the description include physical attributes? & 85\% & 0.71 & 0.76 (0.20)\\
5. Does the description contain hate speech & 1\% & 1 & NA\\
\hline
\end{tabular}
\caption{Questions used in the annotation exercise, with the overall proportion of positive responses across all descriptions. Column 3 gives overall agreement using Fleiss's kappa among the 20 descriptions shared among all 9 participants, including the control annotator. Column 4 gives mean agreement between the 8 annotators and the control annotator, with standard deviations in parentheses. Percentages for questions 2-5 exclude descriptions for which question 1 was answered positively by at least one annotator.}
\label{tab:annot}
\end{table*}

\subsection{Annotation and agreement}
As shown in Figure \ref{fig:example}, there is considerable variation in the descriptions that people write. While the majority include physical features, there are also emotional descriptors, as well as analogical descriptions (as when a person is described as resembling a criminal; see Figure \ref{fig:example_male}) and inferred characteristics (such as the inferred nationality of the man in Figure \ref{fig:example_male}). In addition, such data collection exercises raise a potential ethical concern, insofar as individuals may take advantage of the anonymity of the crowdsourcing exercise to produce texts which are racially or sexually offensive.
We note, however, that while we are taking steps (described below) to identify and weed out such ethically problematic descriptions prior to dissemination and use of the data, we do not intend to exclude descriptions simply on the grounds that they describe nationality or ethnicity, as when a participant described the man in Figure \ref{fig:example_male} as `middle eastern'. Indeed, such examples raise interesting questions about the salience of such features for different individuals, as a function, for instance, of where they come from (and hence, of what counts as `frequent' or `the norm' in their cultural milieu). By including country of origin, age and gender among the demographic details we request from participants, we hope to be able to address these questions in a more informed manner.

\subsection{Annotation}
To gain a better understanding of these issues, as well as potential challenges in annotating the data, we conducted an annotation exercise on a subset of the data. The study was conducted  among all nine members of the RIVAL team. Eight of these were designated as annotators, while a ninth acted as a control annotator.

Each of the 8 annotators was assigned a random set of 194 descriptions. Of these, a random set of 20 descriptions were shared among all eight annotators. This was used to compute overall agreement. The control annotator's data consisted of 180 descriptions consisting of (a) the shared subset of 20 descriptions; (b) a further 160 descriptions, 20 from each of the 8 annotators' set.

All participants viewed each description and its corresponding image, and responded to the set of yes/no questions shown in the left panel of Table \ref{tab:annot}. The first of these was intended to weed out descriptions which have no relationship to the image. Descriptions for which annotators answered this question positively are not included in subsequent analyses. The last question in Table \ref{tab:annot} was intended to identify potentially racist or sexually discriminatory descriptions, modulo the provisions made above concerning the use of ethnic or other characteristics when these are not used in an offensive manner. As shown in Table \ref{tab:annot}, only one description was identified as potentially containing hate speech.

The overall proportions of positive responses suggest that the majority of descriptions focus on physical attributes, as expected, but a substantial proportion also incorporate inferred characteristics and/or emotional elements (such as whether a person looks happy or sad).

The table also gives the agreement between annotators on the shared set of 20 descriptions, estimated using Fleiss's kappa, which is appropriate for multiple annotators. As shown in the table, most questions had high levels of agreement with $\kappa \geq 0.7$. The exception is the second question, where agreement falls to $0.48$. This indicates that what is `inferred' is open to interpretation, with some annotators viewing features such as nationality, and even a person's age, as inferred, while others do not. During discussions between annotators, after the exercise was completed, it became clear that age was sometimes viewed as a physical feature (insofar as it can be deduced from a person's physical characteristics), while others viewed it as an inference.

\subsection{Agreement}
As a sanity check, we also computed agreement between each of the 8 annotators and the control annotator. Recall that, in addition to the set of 20 descriptions shared among all participants, the latter annotated a further 160 descriptions, consisting of 20 from each of the 8 annotators' cases. The agreement results are shown in the final column of the table, with the exception of Question 5, since none of the descriptions shared with the control annotator were classified by any individual participant as containing hate speech. In this case, agreement figures are generally lower, but they are means over 8 distinct values. Perhaps the most notable drop is in question 4, which deals with descriptions containing emotion, where mean agreement with the control annotator drops to 0.49. Note, however, that this is also the case where variance is highest ({\sc sd}=0.33). One important reason for the drop is that, in this case, one annotator seems to have interpreted the question very differently from the control annotator, resulting in a negative agreement figure\footnote{Negative agreement implies less agreement than expected by chance.} and increasing the variance considerably (range: $\langle -0.19, 0.88 \rangle$). The mean without this outlier goes up to 0.60.

\subsection{Discussion}
Overall, the distribution in the sub-corpus included in the annotation study conforms to expectations, with a majority of descriptions incorporating physical attributes, and a sizeable proportion including emotional and possibly inferred attributes.

As for agreement, the annotators appear to have reliably identified descriptions as falling into the five categories of interest, with the possible exception of inference, which clearly needs a more precise definition. Agreement with the control annotator is generally lower across the board,

One of the consequences of this preliminary study is that we are better placed to predict what a face description dataset will contain, and what the challenges for automatic face description will be. In particular, descriptions are bound to refer to a mixture of attributes, not only physical, but also emotional. The latter are probably more challenging to identify with current CV tools, but may also raise interesting questions about how they should be expressed in relation with physical characteristics. Does a person who is smiling qualify as happy?

Clearly, what is expressed will also depend on the purpose which the descriptions are intended to achieve, though the present crowd-sourcing study did not specify a particular purpose, since the aim was to cast the net as wide as possible with a view to gaining a better understanding of the ways in which people describe faces.

\section{Conclusions and Future Work}
This paper described ongoing work on the construction of a large corpus of descriptions of human faces, together with an annotation and agreement study whose purpose was to identify the main distribution of types of descriptions, and the extent to which annotators can expect to agree on these types.

Our current work is focussing on extending the crowd-sourcing study to produce a dataset that is sufficiently large to support non-trivial machine learning work on the automated description of faces. This will extend the reach of current image-to-text systems, to a domain where the focus is necessarily less on scene descriptions involving objects and relations, and more on fine-grained descriptions using physical and other attributes.
Based on the annotation reported in this paper, we intend to filter irrelevant (`garbage') descriptions and/or ethically problematic ones prior to dissemination.

Apart from extending the crowd-sourcing exercise to elicit more human-authored descriptions, we are actively exploring semi-automatic data augmentation techniques. One particularly promising avenue is to use the existing image-description pairs in the Face2Text corpus to harvest similar, publicly available images. This can be done via the web (e.g. using Google's image search service). Our goal is to mine the text surrounding such images to find portions of text that are similar to (portions of) the descriptions produced by the contributors in our corpus.

A second challenge will be to address possible differences in the purposes for which such descriptions can be produced. The possibilities here are very wide-ranging, from describing a face accurately enough for recognition (for instance, in a forensic context where a description is required to construct an identifiable image), to more gamified or humorous contexts, where descriptions might need to rely more on analogy or inference.

In the medium-term, one of our goals is to undertake a more fine-grained annotation exercise on our existing data, with a view to identifying portions of descriptions that pertain to particular features (physical, emotional, ethnic etc). Our agreement statistics already indicate that these can be identified with reasonably high reliability; by explicitly annotating the data, we hope to be able to develop techniques to automatically tag future descriptions with this high-level semantic information. Based on this, it will be possible to undertake a more fine-grained evaluation of the corpus, for example, to find the types of images for which certain attributes tend to recur in people's descriptions. A further aim is to correlate such trends with the demographic data we collect from participants, with a view to constructing a model to predict which aspects of a face will be salient enough to warrant explicit mention, given the describer's own characteristics and background.

\section*{Acknowledgements}
We thank our three anonymous reviewers for helpful comments on the paper.
This work is partially funded by the Endeavour Scholarship Scheme (Malta). Scholarships are part-financed by the European Union - European Social Fund (ESF) - Operational Programme II - Cohesion Policy 2014-2020.

\section{References}
\label{main:ref}

\bibliographystyle{lrec}
\bibliography{rival_lrec}

\begin{thebibliography}{}

\bibitem[\protect\citename{Berg \bgroup et al.\egroup }2005]{berg2005}
Berg, T.~L., Berg, A.~C., Edwards, J., and Forsyth, D.~A.
\newblock (2005).
\newblock Who's in the picture.
\newblock In {\em Advances in neural information processing systems}, pages
  137--144.

\bibitem[\protect\citename{Bernardi \bgroup et al.\egroup
  }2016]{Bernardietal2015}
Bernardi, R., {\c{C}}akici, R., Elliott, D., Erdem, A., Erdem, E.,
  Ikizler{-}Cinbis, N., Keller, F., Muscat, A., and Plank, B.
\newblock (2016).
\newblock Automatic description generation from images: {A} survey of models,
  datasets, and evaluation measures.
\newblock {\em Journal of Artificial Intelligence Research (JAIR)},
  55:409--442.

\bibitem[\protect\citename{Devlin \bgroup et al.\egroup }2015]{Devlin2015}
Devlin, J., Cheng, H., Fang, H., Gupta, S., Deng, L., He, X., Zweig, G., and
  Mitchell, M.
\newblock (2015).
\newblock {Language Models for Image Captioning: The Quirks and What Works}.
\newblock In {\em Annual Meeting of the Association for Computational
  Linguistics}.

\bibitem[\protect\citename{Elliott and Keller}2013]{ElliottKeller2013}
Elliott, D. and Keller, F.
\newblock (2013).
\newblock Image description using visual dependency representations.
\newblock In {\em Proceedings of the 2013 Conference on Empirical Methods in
  Natural Language Processing (EMNLP '13}, pages 1292--1302, Seattle,
  Washington, U.S.A.

\bibitem[\protect\citename{Elliott \bgroup et al.\egroup }2016]{multi30k}
Elliott, D., Frank, S., Sima'an, K., and Specia, L.
\newblock (2016).
\newblock Multi30k: Multilingual english-german image descriptions.
\newblock {\em CoRR}.

\bibitem[\protect\citename{Hodosh \bgroup et al.\egroup }2013]{hodosh2013}
Hodosh, M., Young, P., and Hockenmaier, J.
\newblock (2013).
\newblock {Framing Image Description as a Ranking Task: Data, Models and
  Evaluation Metrics}.
\newblock {\em Journal of Artificial Intelligence Research}, 47:853--899.

\bibitem[\protect\citename{Huang \bgroup et al.\egroup }2007]{LFW}
Huang, G.~B., Ramesh, M., Berg, T., and Learned-Miller, E.
\newblock (2007).
\newblock Labeled faces in the wild: A database for studying face recognition
  in unconstrained environments.
\newblock Technical Report 07--49.

\bibitem[\protect\citename{Kemelmacher-Shlizerman \bgroup et al.\egroup
  }2016]{kemelmacher2016megaface}
Kemelmacher-Shlizerman, I., Seitz, S.~M., Miller, D., and Brossard, E.
\newblock (2016).
\newblock The megaface benchmark: 1 million faces for recognition at scale.
\newblock In {\em Proceedings of the IEEE Conference on Computer Vision and
  Pattern Recognition}, pages 4873--4882.

\bibitem[\protect\citename{Klare \bgroup et al.\egroup }2015]{Klare2015}
Klare, B.~F., Klein, B., Taborsky, E., Blanton, A., Cheney, J., Allen, K.,
  Grother, P., Mah, A., Burge, M., and Jain, A.~K.
\newblock (2015).
\newblock Pushing the frontiers of unconstrained face detection and
  recognition: Iarpa janus benchmark a.
\newblock In {\em 2015 IEEE Conference on Computer Vision and Pattern
  Recognition (CVPR)}, pages 1931--1939.

\bibitem[\protect\citename{Kulkarni \bgroup et al.\egroup }2011]{Kulkarni2011}
Kulkarni, G., Premraj, V., Dhar, S., Li, S., Choi, Y., Berg, A.~C., and Berg,
  T.~L.
\newblock (2011).
\newblock {Baby talk: Understanding and generating simple image descriptions}.
\newblock In {\em IEEE Conference on Computer Vision and Pattern Recognition}.

\bibitem[\protect\citename{Kuznetsova \bgroup et al.\egroup
  }2012]{Kuznetsova2012}
Kuznetsova, P., Ordonez, V., Berg, A.~C., Berg, T.~L., and Choi, Y.
\newblock (2012).
\newblock {Collective Generation of Natural Image Descriptions}.
\newblock In {\em Annual Meeting of the Association for Computational
  Linguistics}.

\bibitem[\protect\citename{L. \bgroup et al.\egroup
  }2015]{liu2015faceattributes}
L., Z., L., P., W., X., and T., X.
\newblock (2015).
\newblock Deep learning face attributes in the wild.
\newblock In {\em Proceedings of International Conference on Computer Vision
  (ICCV)}.

\bibitem[\protect\citename{Learned-Miller \bgroup et al.\egroup
  }2016]{Learned-Miller2016}
Learned-Miller, E., Huang, G.~B., RoyChowdhury, A., Li, H., and Hua, G.
\newblock (2016).
\newblock Labeled faces in the wild: A survey.
\newblock In M.~Kawulok, et~al., editors, {\em Advances in Face Detection and
  Facial Image Analysis}, pages 189--248. Springer International Publishing,
  Berlin and Heidelberg.

\bibitem[\protect\citename{Lin \bgroup et al.\egroup }2014]{MSCOCO}
Lin, T., M., M., Belongie, S.~J., Bourdev, L.~D., Girshick, R.~B., Hays, J.,
  Perona, P., Ramanan, D., Doll{\'{a}}r, P., and Zitnick, C.~L.
\newblock (2014).
\newblock Microsoft {COCO:} common objects in context.
\newblock {\em CoRR}, abs/1405.0312.

\bibitem[\protect\citename{Mao \bgroup et al.\egroup }2014]{Mao2015}
Mao, J., Xu, W., Yang, Y., Wang, J., Huang, Z., and Yuille, A.
\newblock (2014).
\newblock Deep captioning with multimodal recurrent neural networks (m-rnn).
\newblock {\em {ICLR} 2015}.

\bibitem[\protect\citename{Mitchell \bgroup et al.\egroup }2012]{Mitchell2012}
Mitchell, M., Dodge, J., Goyal, A., Yamaguchi, K., Stratos, K., Han, X.,
  Mensch, A., Berg, A., Han, X., Berg, T., and {Daume III}, H.
\newblock (2012).
\newblock {Midge: Generating Image Descriptions From Computer Vision
  Detections}.
\newblock In {\em Proceedings of the 13th Conference of the European Chapter of
  the Association for Computational Linguistics (EACL'12)}, pages 747--756,
  Avignon, France. Association for Computational Linguistics.

\bibitem[\protect\citename{Ordonez \bgroup et al.\egroup }2011]{Ordonez2011}
Ordonez, V., Kulkarni, G., and Berg, T.~L.
\newblock (2011).
\newblock {Im2text: Describing images using 1 million captioned photographs}.
\newblock In {\em Advances in Neural Information Processing Systems}.

\bibitem[\protect\citename{Reed \bgroup et al.\egroup }2016]{reed2016}
Reed, S., Akata, Z., Schiele, B., and Lee, H.
\newblock (2016).
\newblock Learning deep representations of fine-grained visual descriptions.
\newblock In {\em IEEE Computer Vision and Pattern Recognition}.

\bibitem[\protect\citename{Rennie \bgroup et al.\egroup }2017]{Rennie2016}
Rennie, S.~J., Marcheret, E., Mroueh, Y., Ross, J., and Goel, V.
\newblock (2017).
\newblock Self-critical sequence training for image captioning.
\newblock In {\em Proceedings of the Ieee Conference on Computer Vision and
  Pattern Recognition (cvpr)}.

\bibitem[\protect\citename{Socher \bgroup et al.\egroup }2014]{Socher2014}
Socher, R., Karpathy, A., Le, Q.~V., Manning, C.~D., and Ng, A.
\newblock (2014).
\newblock {Grounded Compositional Semantics for Finding and Describing Images
  with Sentences}.
\newblock {\em Transactions of the Association for Computational Linguistics},
  2:207--218.

\bibitem[\protect\citename{Tanti \bgroup et al.\egroup }2018]{Tanti2018}
Tanti, M., Gatt, A., and Camilleri, K.~P.
\newblock (2018).
\newblock Where to put the image in an image caption generator.
\newblock {\em CoRR}, abs/1703.09137.

\bibitem[\protect\citename{Tottenham \bgroup et al.\egroup }2009]{NimStim}
Tottenham, N., Tanaka, J.~W., Leon, A.~C., McCarry, T., Nurse, M., Hare, T.~A.,
  J., M.~D., Westerlund, A., Casey, B., and Nelson, C.
\newblock (2009).
\newblock The nimstim set of facial expressions: Judgments from untrained
  research participants.
\newblock {\em Psychiatry research}, 168(3):242--249.

\bibitem[\protect\citename{Vinyals \bgroup et al.\egroup }2015]{Vinyals2015}
Vinyals, O., Toshev, A., Bengio, S., and Erhan, D.
\newblock (2015).
\newblock Show and tell: A neural image caption generator.
\newblock In {\em Proceedings of the {IEEE} Conference on Computer Vision and
  Pattern Recognition ({CVPR})}, pages 3156--3164.

\bibitem[\protect\citename{Xu \bgroup et al.\egroup }2015]{Xu2015}
Xu, K., Ba, J., Kiros, R., Cho, K., Courville, A., Salakhudinov, R., Zemel, R.,
  and Bengio, Y.
\newblock (2015).
\newblock Show, attend and tell: Neural image caption generation with visual
  attention.
\newblock volume~37 of {\em Proceedings of Machine Learning Research}, pages
  2048--2057, Lille, France. Pmlr.

\bibitem[\protect\citename{Yin \bgroup et al.\egroup }2006]{yin2006}
Yin, L., Wei, X., Sun, Y., Wang, J., and Rosato, M.~J.
\newblock (2006).
\newblock A 3d facial expression database for facial behavior research.
\newblock In {\em Automatic face and gesture recognition, 2006. FGR 2006. 7th
  international conference on}, pages 211--216. IEEE.

\bibitem[\protect\citename{Young \bgroup et al.\egroup }2014]{young2014}
Young, P., Lai, A., Hodosh, M., and Hockenmaier, J.
\newblock (2014).
\newblock From image descriptions to visual denotations: New similarity metrics
  for semantic inference over event descriptions.
\newblock {\em Transactions of the Association for Computational Linguistics},
  2:67--78.

\end{thebibliography}


\end{document}